\documentclass{article}

\usepackage{PRIMEarxiv}

\usepackage[utf8]{inputenc} 
\usepackage[T1]{fontenc}    
\usepackage{hyperref}       
\usepackage{url}            
\usepackage{booktabs}       
\usepackage{amsfonts}       
\usepackage{nicefrac}       
\usepackage{microtype}      
\usepackage{lipsum}
\usepackage{fancyhdr}       
\usepackage{graphicx}       
\graphicspath{{media/}}     

\usepackage[backend=biber, style=ieee, dashed=false]{biblatex}

\title{Human interaction classifier for LLM based chatbot}

\author{
  Diego Martín, Jordi Sanchez, Xavier Vizcaíno \\
  Applus IDIADA \\
  Santa Oliva\\
  \texttt{\{diego.martin, jordi.sanchez, xavier.vizcaino\}@idiada.com} \\
}

\begin{document}
\maketitle

\begin{abstract}

This study investigates different approaches to classify human interactions in an artificial intelligence-based environment, specifically for Applus+ IDIADA's intelligent agent AIDA. The main objective is to develop a classifier that accurately identifies the type of interaction received (Conversation, Services, or Document Translation) to direct requests to the appropriate channel and provide a more specialized and efficient service. Various models are compared, including LLM-based classifiers, KNN using Titan and Cohere embeddings, SVM, and artificial neural networks. Results show that SVM and ANN models with Cohere embeddings achieve the best overall performance, with superior F1 scores and faster execution times compared to LLM-based approaches. The study concludes that the SVM model with Cohere embeddings is the most suitable option for classifying human interactions in the AIDA environment, offering an optimal balance between accuracy and computational efficiency. 

\end{abstract}

\section*{Acknowledgements}
To Karim Belaid, to review and help us make clear movements towards this paper submission. Your expertise and insights will be invaluable in refining our research, strengthening our arguments, and ensuring the quality of our work meets the highest standards. We greatly appreciate your time and dedication in supporting this important academic endeavor.

\section{Introduction}

Artificial intelligence has emerged as a transformative force, reshaping the landscape of modern society. Its influence extends to various sectors, including the automotive industry where Applus+ IDIADA operates. Within this environment, AI has taken the form of an intelligent chatbot baptized as AIDA, an LLM-powered virtual assistant.

An LLM-based chatbot is an artificial intelligence system that uses a Large Language Model (LLM) to generate human-like responses in conversations. These chatbots are trained on vast amounts of text data, allowing them to understand and produce natural language across a wide range of topics. They can engage in more complex, context-aware dialogues compared to traditional rule-based chatbots, often demonstrating impressive language understanding and generation capabilities. 

This digital aide lends its capabilities to a multitude of tasks, serving as a versatile companion to IDIADA's workforce. From addressing inquiries to tackling intricate technical challenges spanning code, mathematics, and translation, AIDA's versatility knows no bounds.

At a certain point, the need to establish a clear criterion for the categorization of the various interactions that this intelligent agent receives on a daily basis arises. The main aim is to offer a more in-depth service through the creation of dedicated pipelines for various contexts (conversation, document translation, services).

By categorizing the interactions into these three main groups, the system can better understand the user's intent and respond accordingly. The Conversation class encompasses general inquiries and exchanges, the Services class covers requests for specific functionalities or support, and the Document Translation class handles text translation needs.

In this paper, we will discuss the research process carried out to develop a classifier for human interactions in this AI-based environment. The objective of this classifier is to accurately identify the type of interaction received by the intelligent agent AIDA, in order to route the request to the appropriate pipeline and provide a more specialized and efficient service.

Through this classification model, the AIDA assistant can optimize its responses, allocate resources more effectively, and enhance the overall user experience. The research and development of this LLM-based classifier is an important step in the continuous improvement of the intelligent agent's capabilities within the Applus IDIADA ecosystem.

\section{The data}

We are going to use a set of 1668 examples of pre-classified human interactions. These will be divided into 666 for training and 1002 for testing. A 40-60 has been applied, thus giving a lot of importance to the test set.

The following figure shows some of the examples:

\begin{figure}[htp]
    \centering
    \includegraphics[width=15cm]{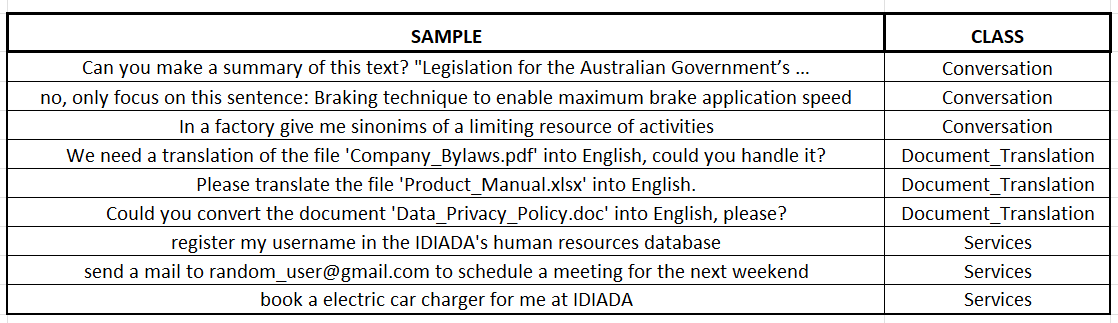}
    \caption{Human Interaction Samples}
    \label{fig:examples}
\end{figure}

\hspace{1cm}

All samples have been gathered from real AIDA users interactions. All the samples remain in a RAW state, unadulterated and without preprocess. They have been manually labeled by the Datalab team. There are no mislabeled cases, despite the complexity in some cases given the abstractness of the interaction itself.

\begin{figure}[htp]
    \centering
    \includegraphics[width=10cm]{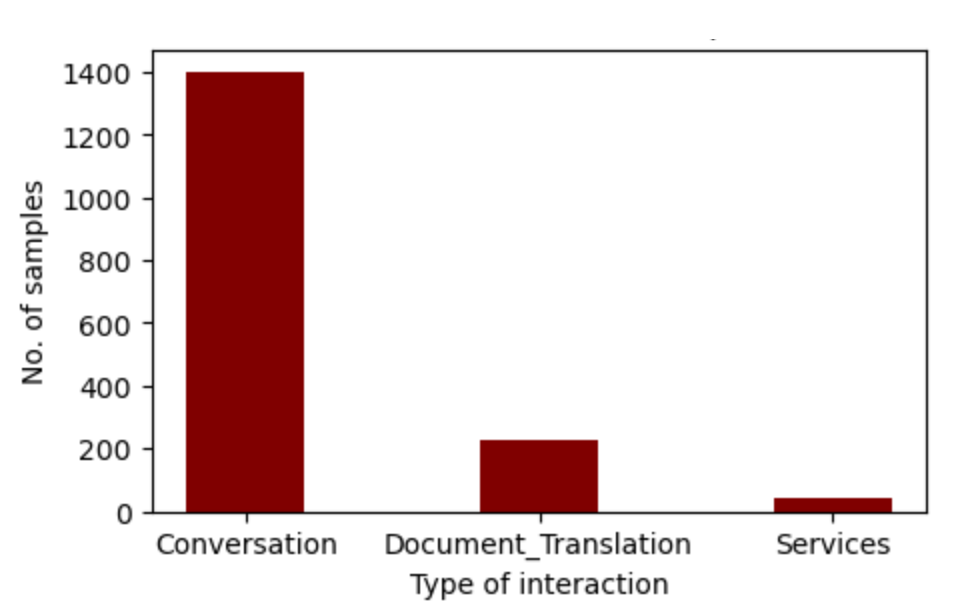}
    \caption{Distribution of examples}
    \label{fig:examples_distribution}
\end{figure}

\newpage
\section{Methodology}

In this section three different methods are presented, two of them based in LLM models and another one based in machine learning classic algorithm. The aim is to understand which approach it is the most suitable to face the presented problem.

\textbf{NOTE: in all cases the prompt or the implementation can be seen in the Annex section.\\}

\subsection{LLM Based Classifier: Simple prompt}

In this case, we present an LLM-based classifier to discern between three classes: Conversation, Services, and Document Translation. This initial approach is based on the principle of understanding a set, through which the nature of the cases that compose it is defined.

Therefore, three different definitions have been given to the model, one for each one of the classes (Conversation, Services, Document\_translation). It is important to note that maybe, the definition that It could be expected for some class, It is not exactly the one It could be held in mind. For instance, in case of Conversation, we can see that we englobe tasks about text summarisation.

\subsection{LLM Based Classifier: Example Augmented Inference}

This approach uses RAG (retrieval augmented generation) techniques to feed the model response, not only with a definition by compression, but also a quasi-definition by extension. That is, in this case we show the model different examples of each class with the aim that the model is able to learn the inherent characteristics of each one in order to complement the initial description given.

\begin{figure}[htp]
    \centering
    \includegraphics[width=8cm]{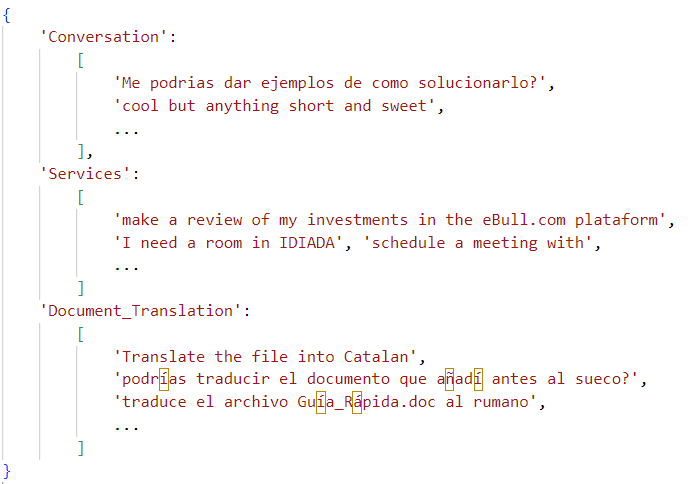}
    \caption{Examples in json format for RAG}
    \label{fig:examples_json}
\end{figure}

The total amount of examples given for each class is:

 \begin{itemize}
        \item \textbf{Conversation}: 500 examples. This is the most common class, only 500 samples are given to the model given the Huge amount of information held which causes overflow of the infrastructure (very high delays, throttling, connections shutouts).

        This is something very important to note, because it is a big stopper. Giving more examples this approach could perform better, but the question is ¿How many examples? Surely a huge amount of them. 
            
        \item \textbf{Services}: 26 examples. This is the less common class, In this case all train data have been used.

        \item \textbf{Translate\_Document}: 140 examples. In this case all train data have been used.
\end{itemize}

This approach encounters a challenge absent in the previous one: Scalability. While the model's performance benefits from additional examples, the mounting computational demands render it incompatible with the current infrastructure. The sheer volume of data becomes unwieldy, leading to quota issues with Bedrock services and very long time responses. This is not acceptable, there is a need of rapid response times to maintain a satisfactory user experience.

\newpage
\subsection{KNN Based Classifier: Titan Embeddings}

On this occasion we approach the problem taking into account that despite the multitude of different interactions that may exist, at the end of the day these are very repeatable or similar to each other. So, perhaps, when making a new estimation for an input, a distance-based approach could be applied, such as KNN (K-NEAREST-NEIGHBORS) \cite{fix1951discriminatory}, assigning the class of the most similar examples surrounding the input.

To achieve this, it is imperative to transform the interactions, which are in textual form, into a space where algebraic operations can be applied. To this end,a technique known as Embeddings has been applied, which facilitates the derivation of a coordinate vector for a text within a space. Embeddings are vector representations of text that capture semantic and contextual information. Consequently, by comparing two vectors, the semantic similarity between two different interactions is analyzed, thereby determining "their proximity."

\begin{itemize}

    \item \textbf{Embeddings}: Amazon Titan 1 has been used. This model generate vectors of 1536 dimensions, it is trained to accept several languages and to retain the semantic meaning of the phrases embedded.
    
    \item \textbf{Classifier}: a classic machine learning algorithm (knn) has been applied by using the sklearn python module. This method takes a parameter, which has been set to 3.

    \begin{figure}[htp]
        \centering
        \includegraphics[width=6cm]{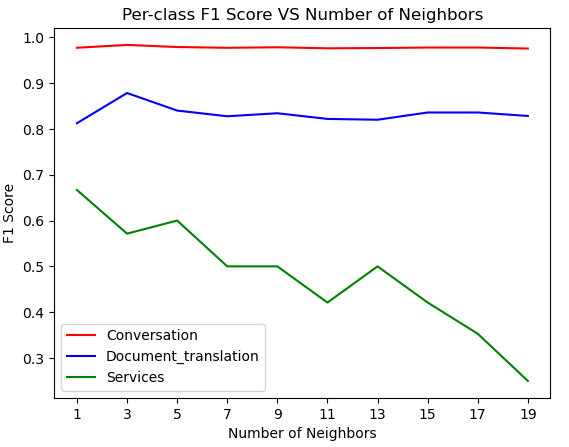}
        \caption{Per-class F1 Score VS Number of Neighbors}
        \label{fig:codo}
    \end{figure}

    This figure shows the F1 scores for each one of the classes vs the Number of neighbors used. As it shows, the best point is k=3 (highest value for Document\_Translation, 3rd highest value for Services, but Document\_Translation is much more common than Services).

\end{itemize}

\hspace{1cm}
\subsection{KNN Based Classifier: Cohere Embeddings}

In the previous section, we used the Amazon Titan 1 model to generate text embeddings due to its popularity. However, there are other models available in the market. In this section, we repeat the same process but using the COHERE multilingual model. We chose this model because of its excellent ability to work with various languages without affecting the vectorization of phrases. As we will demonstrate shortly, this model does not cause significant differences in the generated vectors, thus making it more suitable for use in a multilingual environment such as AIDA.

\begin{itemize}
    \item \textbf{Embeddings}: COHERE model has been used. This model generate vectors of 1024 dimensions, it is trained to accept several languages and to retain the semantic meaning of the phrases embedded.

    \item \textbf{Classifier}: a classic machine learning algorithm (knn) has been applied by using the sklearn python module. This method takes a parameter, which has been set to 11.
    \newpage
    \begin{figure}[htp]
        \centering
        \includegraphics[width=6cm]{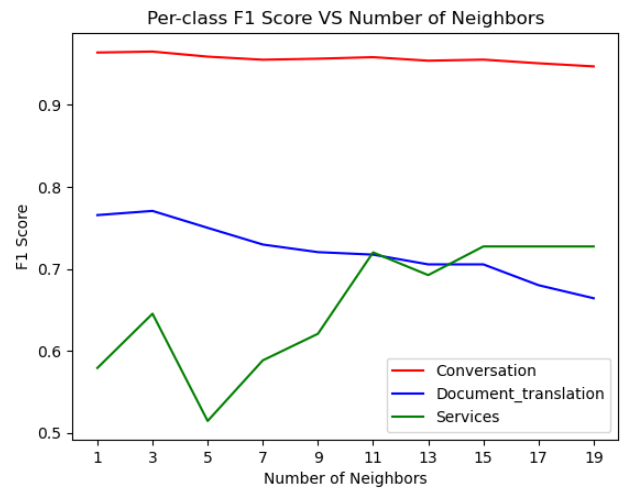}
        \caption{Per-class F1 Score VS Number of Neighbors}
        \label{fig:codo2}
    \end{figure}

    This figure shows the F1 scores for each one of the classes vs the Number of neighbors used. As it shows, the best point is k=11 (highest value for Document\_Translation and 2nd for Services). This time the Documents\_Translation / Services trade-off worth's it.

\end{itemize}

\subsection{Titan Embeddings VS Cohere Embeddings}

In this section we go a little deeper into the embeddings generated by both models, with the aim of understanding their nature and thus being able to understand the results obtained with the models. To achieve this, a dimensionality reduction has been carried out with the aim of being able to visualize the vectors obtained in both cases in 2D.

One point to keep in mind is that the cohere model has a limitation on the size of the text that it is capable of vectorizing, this being an important limit. Therefore, in the implementation shown in the previous section you can see the application of a filter to only have interactions of 1500 characters (which leaves out all cases that exceed the limit).

In the next figure, the vector spaces generated in each case can be observed:

    \begin{figure}[htp]
        \centering
        \includegraphics[width=15cm]{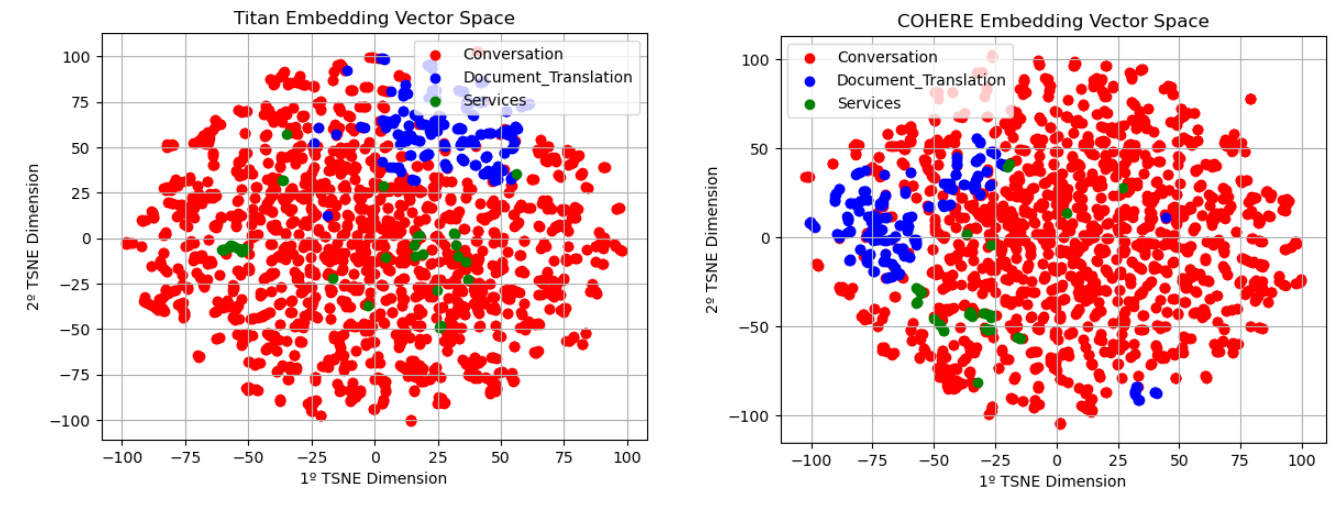}
        \caption{Titan VS Cohere Embedding}
    \end{figure}

As we can see, the spaces generated are relatively similar, it could seem that they are very similar spaces with a rotation between one and the other but looking closely it can be seen that the direction of maximum variance in the case of cohere is different (it can be deduced through from observing the relative position/shape of the different groups).

This type of situation, in which a high class overlap is observed, is a good case for applying algorithms such as KNN. As said in the introduction "most human interactions with AI, at the end of the day, are very similar to each other of the same class" This would explain why KNN-based models outperform LLM-based models.

\newpage

\subsection{SVM Based Classifier: Titan Embeddings}

In this scenario, it is likely that user interactions belonging to the three main categories (Conversation, Services, and Document Translation) form distinct clusters or groups within the embedding space. This is because each category has particular linguistic and semantic characteristics that would be reflected in the geometric structure of the embedding vectors. Remember that in the previous visualization of the embeddings space only 2D transformation of this space is displayed. This does not mean that clusters could not be highly separable in a higher dimension.

Classification algorithms like SVMs \cite{cortes1995support} are especially well-suited to leverage this implicit geometry of the data. SVMs seek to find the optimal hyperplane that separates the different groups or classes in the embedding space, maximizing the margin between them. This ability of SVMs to exploit the underlying geometric structure of the data makes them an intriguing option for this user interaction classification problem.

Furthermore, SVMs are a robust and efficient algorithm that can effectively handle high-dimensional data sets, such as text embeddings. This makes them particularly suitable for this scenario, where the embedding vectors of the user interactions are expected to have a high dimensionality.

\begin{itemize}
    \item \textbf{Embeddings}: Amazon Titan 1 \cite{amazon_titan} has been used.  This model generate vectors of 1536 dimensions, it is trained to accept several languages and to retain the semantic meaning of the phrases embedded.

    \item \textbf{Classifier}: a classic machine learning algorithm (SVM) has been applied by using the sklearn python module. This method takes several parameters, in order to get the optimal ones a grid search with 10-fold cross validation based in f1 multi class score has been applied resulting in the following parameters selected:

    \begin{itemize}
        \item \textbf{C}: 1. This parameter controls the trade-off between allowing training errors and forcing rigid margins. It is a regularization parameter. A higher value of C (e.g., 10) indicates a higher penalty for miss-classification errors, resulting in a more complex model that tries to fit the training data more closely. This can be useful when the classes in the data are well-separated, as it allows the algorithm to create a more intricate decision boundary to accurately classify the samples. A C of 1 indicates a reasonable balance between fitting the training set and the model's generalization ability which could mean the data has a simple structure and that a more flexible model is not necessary to capture the underlying relationships.
        
        \item \textbf{class\_weight}: None. This parameter adjusts the weights of each class during the training process. Setting 'class\_weight' to 'balanced' automatically adjusts the weights inversely proportional to the class frequencies in the input data. This is particularly useful when dealing with imbalanced datasets, where one class is significantly more prevalent than the other(s). A value of None suggest that the minnor clases do not have much relevance which in the same manner suggest that the implicit geometry not be good enough in order to separate the different classes.
        
        \item \textbf{kernel}: 'linear'. This parameter specifies the type of kernel function to be used by the SVC algorithm. The 'linear' kernel is a simple and efficient choice, as it assumes that the decision boundary between classes can be represented by a linear hyperplane in the feature space. This value suggest that in a higher dimension vector space the categories could be linearly separated by an hyperplane.
    \end{itemize}
\end{itemize}

\subsection{SVM Based Classifier: Cohere Embeddings}

\begin{itemize}
    \item \textbf{Embeddings}: COHERE \cite{cohere_embeddings} model has been used. This model generate vectors of 1024 dimensions, it is trained to accept several languages and to retain the semantic meaning of the phrases embedded.
    
    \item \textbf{Classifier}: a classic machine learning algorithm (SVM) has been applied by using the sklearn python module. This method takes several parameters, in order to get the optimal ones a grid search with 10-fold cross validation has been applied resulting in the following parameters selected:

    \begin{itemize}
        \item \textbf{C}: 10. A C of 10 may suggest that the data has a more complex structure and that a more flexible model is necessary to capture the underlying relationships in the data.
        
        \item \textbf{class\_weight}: 'balanced'.In this case 'balanced' suggest that the minor classes may have more relevance than in the previous case which in the same manner suggest that the implicit geometry may be better in order to separate the different classes.
        
        \item \textbf{kernel}: 'linear'. This parameter suggest that in a higher dimension vector space the categories can be successfully linearly separated by an hyperplane.
    \end{itemize}
\end{itemize}

\newpage
\subsection{ANN Based Classifier: Titan/Cohere Embeddings}

We have seen in the final results that SVM can provide good solutions, and therefore we decided to explore another method based on geometry. An Artificial Neural Network (ANN) \cite{goodfellow2016deep} approach has been applied.

In this case, a normalization of the input vectors has been performed to obtain the advantages of normalization when using neural networks. The normalization of the input data is a crucial step when working with ANNs, as it can help improve the performance and stability of the model. By ensuring that the input features are on a similar scale, the neural network can learn more effectively and avoid issues such as vanishing or exploding gradients during training. A min-max scaling has been applied.

The use of an ANN-based approach provides the ability to capture complex non-linear relationships in the data, which may not be easily modeled using traditional linear methods like SVM. The combination of the geometric insights and the normalization of inputs can potentially lead to improved predictive accuracy compared to the previous SVM results.

\begin{itemize} 
    \item \textbf{Model Definition}: A sequential deep learning model is defined using the Keras library from TensorFlow. \item \textbf{Model Architecture}: The model consists of three densely connected layers. The first layer has 16 neurons and uses the ReLU activation function. The second layer has 8 neurons and also uses the ReLU activation function. The third layer has 3 neurons and uses the softmax activation function. 
    \item \textbf{Model Compilation}: The model is compiled using the categorical\_crossentropy loss function, the Adam optimizer with a learning rate of 0.01, and the categorical\_accuracy metric. An EarlyStopping callback is used to stop the training if the categorical\_accuracy metric does not improve for 25 epochs. 
    \item \textbf{Model Training}: The model is trained for a maximum of 500 epochs using the training set and validated on the test set. The batch size is set to 64. The performance metric used is the maximum classification accuracy (categorical\_accuracy) obtained during the training. 
\end{itemize}

It is important to note that in order to avoid ordinal assumptions an OHE (one hot encoding) representation have been chosen for the output of the net. One-Hot Encoding does not make any assumptions about the inherent order or hierarchy among the categories. This is particularly useful when the categorical variable does not have a clear ordinal relationship, as the model can learn the relationships without being biased by any assumed ordering.

\newpage
\section{Results}

\begin{figure}[htp]
    \centering
    \includegraphics[width=15cm]{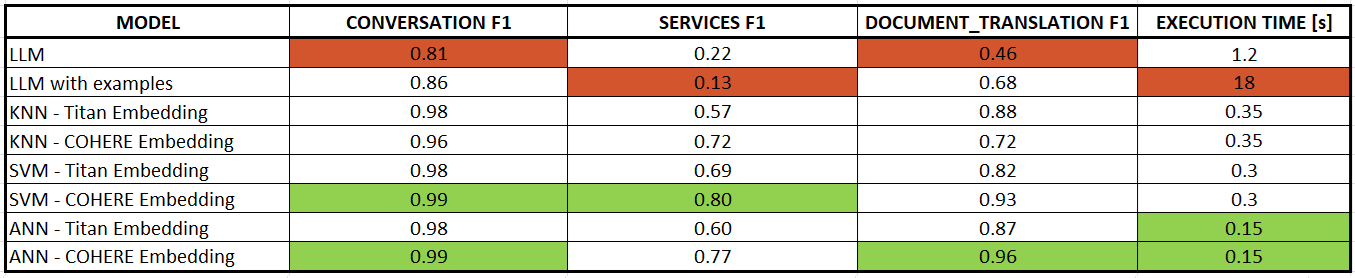}
    \caption{Results}
    \label{fig:results}
\end{figure}

A comparative analysis have been conduced using the code and the data previously presented. The models were assessed based on their F1 scores for the Conversation, Services, and Document Translation tasks, as well as their execution times. This procedure has been conduced several times in order to ensure the results by averaging the obtained metrics for each one of the approaches.

As shown in Figure \ref{fig:results}, the SVM and ANN models leveraging COHERE Embeddings demonstrated the strongest overall performance. The SVM with COHERE Embeddings achieved the highest F1 scores in two out of three tasks, reaching 0.99 for Conversation, 0.80 for Services, and 0.93 for Document Translation. Similarly, the ANN with COHERE Embeddings also performed exceptionally well, with F1 scores of 0.99, 0.77, and 0.96 for the respective tasks.

In contrast, the LLM model exhibited relatively lower F1 scores, particularly for the Services (0.22) and Document Translation (0.46) tasks. However, the performance of the LLM improved when provided with examples, with the F1 score for Document Translation increasing from 0.46 to 0.68.

Regarding execution time, the KNN, SVM, and ANN models demonstrated significantly faster inference times compared to the LLM. The KNN and SVM models with both Titan and COHERE Embeddings had execution times around 0.3-0.35 seconds, while the ANN models were even faster, with execution times of approximately 0.15 seconds. In contrast, the LLM required approximately 1.2 seconds for inference, and the LLM with examples took around 18 seconds.

These results suggest that the SVM and ANN models leveraging COHERE Embeddings offer the best balance of performance and efficiency for the given tasks. The superior F1 scores of these models, coupled with their faster execution times, make them promising candidates for application. The potential benefits of providing examples to the LLM model are also noteworthy, as this approach can help improve its performance on specific tasks.

\section{Conclusions}

In this study, we explore different approaches to classify human interactions in an AI-based environment. Our main objective was to develop a classifier that accurately identifies the type of interaction received by the AIDA intelligent agent, to direct the request to the appropriate channel and provide a more specialized and efficient service.

The comparative analysis of various models conducted in this study has yielded valuable insights into the performance and efficiency trade-offs for the given tasks. The key findings and conclusions are as follows:

\begin{itemize} 
    \item The SVM and ANN models leveraging COHERE Embeddings demonstrated the strongest overall performance, achieving the highest F1 scores across the Conversation, Services, and Document Translation tasks. 
    
    \item The SVM with COHERE Embeddings achieved exceptional results, with F1 scores of 0.99, 0.80, and 0.93 for the respective tasks, making it a promising candidate for real-world applications. 
     \item The ANN model with COHERE Embeddings also exhibited excellent performance, with F1 scores of 0.99, 0.77, and 0.96, further reinforcing the benefits of the COHERE Embedding approach. But this models introduce the need of preprocessing the input vectors in order to normalize them and it may introduce a bias (which min/max values must be used?).
    \item While the LLM model exhibited relatively lower F1 scores, particularly for the Services and Document Translation tasks, providing examples to the LLM was shown to improve its performance on the Document Translation task, increasing the F1 score from 0.46 to 0.68. 
    \item In terms of execution time, the KNN, SVM, and ANN models demonstrated significantly faster inference times compared to the LLM, with execution times around 0.3-0.35 seconds for the KNN and SVM models, and approximately 0.15 seconds for the ANN models. 
    \item The superior performance and efficiency of the SVM and ANN models with COHERE Embeddings make them promising choices for real-world applications that require both high accuracy and low latency. 
\end{itemize}

These findings contribute to the understanding of the trade-offs between model performance, task-specific capabilities, and computational efficiency. The insights gained from this study can inform the selection of appropriate models and preprocessing techniques for various applications, ultimately enhancing the effectiveness and practicality of the deployed systems.

The findings of this study clearly demonstrate the superior performance and efficiency of the SVM model with COHERE Embeddings, making it the most suitable choice for the classification of human interactions in the AIDA intelligent agent environment. The SVM model's exceptional F1 scores across the Conversation, Services, and Document Translation tasks, coupled with its fast inference time, position it as a highly promising candidate for real-world applications.

The ability of the SVM with COHERE Embeddings to accurately identify the type of interaction and direct the request to the appropriate channel will enable AIDA to provide a more specialized and efficient service, ultimately enhancing the overall user experience. Furthermore, the insights gained from this comparative analysis can guide the selection of appropriate models and preprocessing techniques for similar classification tasks, contributing to the development of more effective and practical AI-based systems.

\newpage
\section{Annex}
\subsection{Simple prompt reproducibility}
\hspace{1cm}
\begin{itemize}

    \item \textbf{Libraries}: The programming language employed in this code is Python, complemented by the Langchain module, which is specifically designed to facilitate the integration and utilization of large language models (LLMs). This module provides a comprehensive set of tools and abstractions that streamline the process of incorporating and deploying these advanced artificial intelligence models.\\ 

    To leverage the power of these language models, the Amazon Bedrock service has been utilized. Bedrock is a managed service offered by Amazon Web Services (AWS) that is specifically tailored for the deployment and scaling of large language models. The integration with Bedrock is achieved through the boto3 Python module, which serves as an interface to the AWS ecosystem, enabling seamless interaction with the Bedrock service and the deployment of the classification model.\\

    \item \textbf{Prompt}: The task at hand is to assign one of the three classes (Conversation, Services, or Document\_Translation) to a given sentence, represented by {question}.
    \begin{itemize}
        \item \textbf{Conversation Class}: This class encompasses casual messages, summarization requests, general questions, affirmations, greetings, and similar types of text. It also includes requests for text translation, text summarization, or explicit inquiries about the meaning of words or sentences in a specific language.
        \item \textbf{Services Class}: Texts belonging to this class consist of explicit requests for services such as room reservations, hotel bookings, dining services, cinema information, tourism-related inquiries, and similar service-oriented requests.
        \item \textbf{Document\_Translation Class}: This class is characterized by requests for the translation of a document to a specific language. Unlike the Conversation class, these requests do not involve summarization. Additionally, the name of the document to be translated and the target language are specified.\\ 
    \end{itemize}
    The prompt suggests a hierarchical approach to the classification process. First, the sentence should be evaluated to determine if it can be classified as a Conversation. If the sentence does not fit the Conversation class, then one of the other two classes (Services or Document\_Translation) should be assigned.\\ 

    The priority for Conversation class yields in the fact that \textbf{the 99\% of the interactions are actually simple questions regarding some matters.\\ 
    }

    \item \textbf{Model invocation}: 
        \begin{itemize}
            \item \textbf{Model}: Anthropic Claude 3 Sonnet model has been used \cite{anthropic_claude}, which it is the language model to be used for the natural language processing task. This LLM model has a context window of 200.000 tokens, it is able to manage different languages and to retrieve highly accurate answers.
            \item \textbf{Parameters: max\_tokens}: This parameter limits the maximum number of tokens (words or subwords) that the language model can generate in its output to 50.
            \item \textbf{Parameters: temperature}: This parameter controls the randomness of the language model's output. A temperature of 0.0 means that the model will produce the most likely output according to its training, without any randomness introduced.\\
            
        \end{itemize}

    \item \textbf{Output parser}: another important part is the output parser, which allow us to gather the desired information in JSON format. In order to achieve that, \textbf{langchain output\_parsers} has been used.
\end{itemize}
\newpage
\begin{figure}[htp]
    \centering
    \includegraphics[width=18cm]{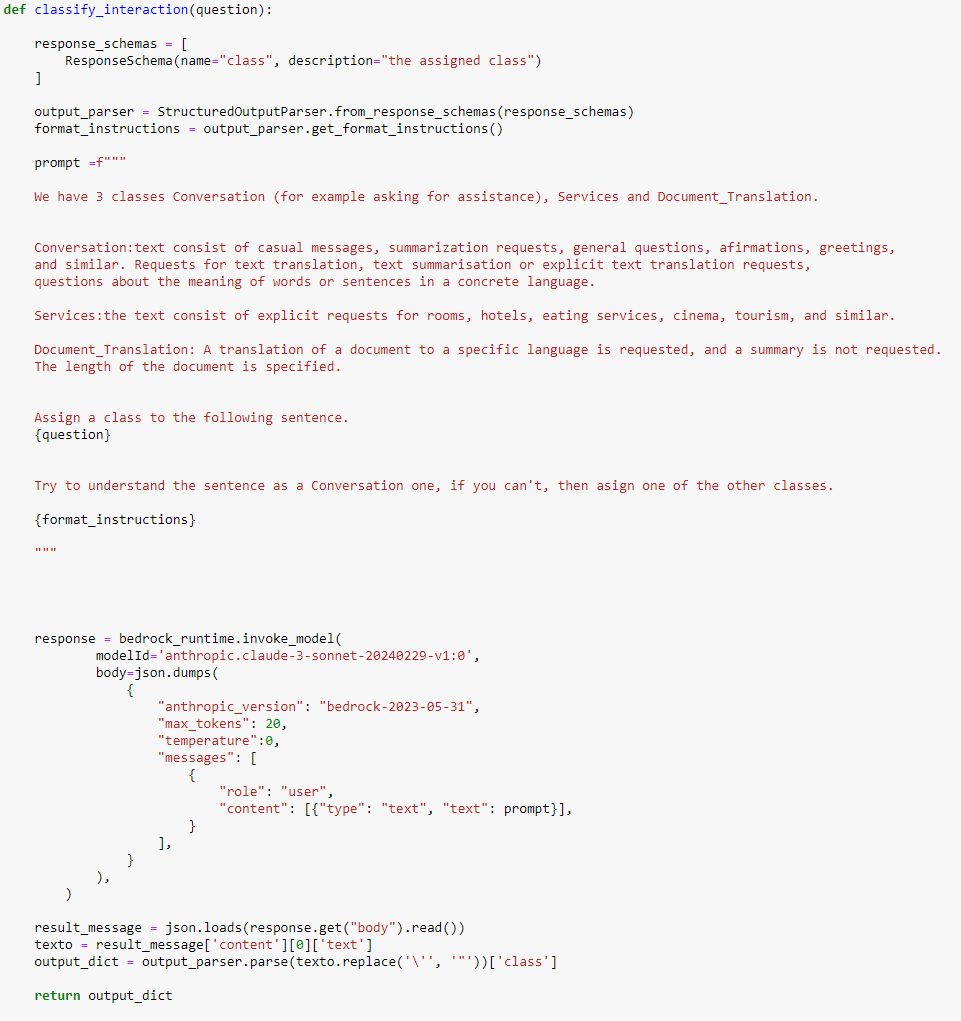}
    \caption{Simple prompt approach}
    \label{fig:simple_prompt}
\end{figure}

\newpage

\subsection{Example Augmented Prompt Inference reproducibility}
\hspace{1cm}
\begin{itemize}
    \item \textbf{Prompt}: in this case the prompt is modified to include all the examples in json format under "Here you have some examples" section.
\end{itemize}

The next figure shows the changes applied to the first version of the classifier.

\begin{figure}[htp]
    \centering
    \includegraphics[width=17cm]{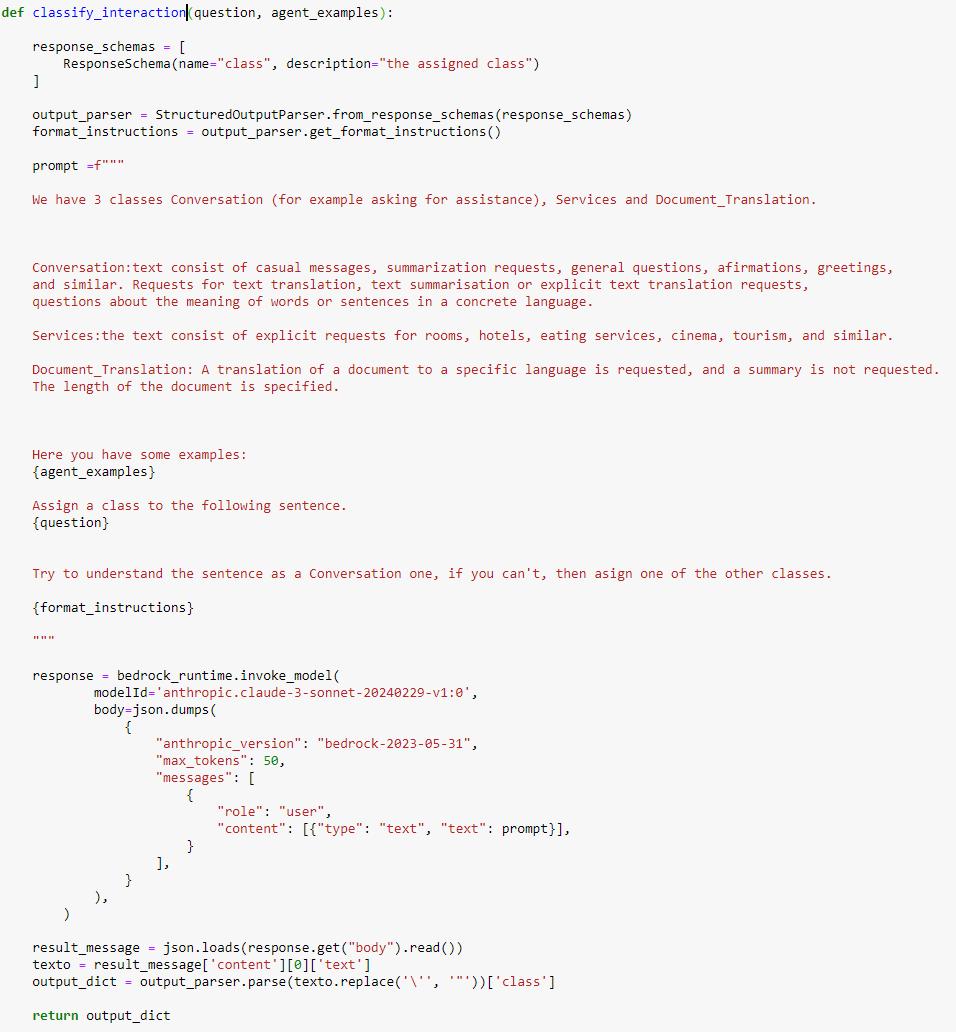}
    \caption{Examples Augmented prompt approach}
    \label{fig:augmented_prompt}
\end{figure}

\subsection{KNN Based Classifier: Titan Embeddings reproducibility}

In the next figure the implementation is presented:

\begin{figure}[htp]
    \centering
    \includegraphics[width=10cm]{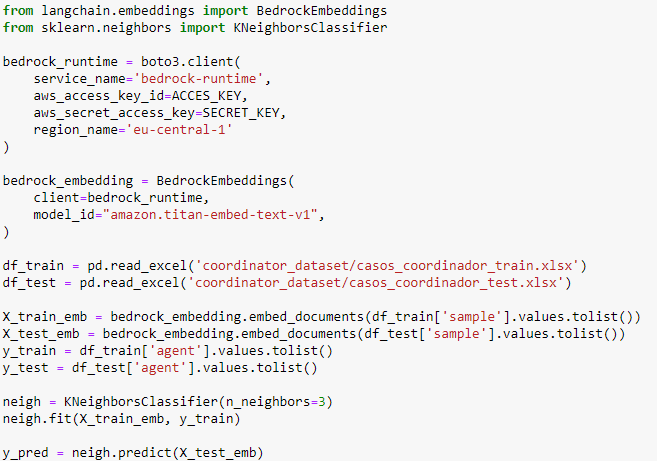}
    \caption{KNN similarity approach}
    \label{fig:knn1}
\end{figure}

\subsection{KNN Based Classifier: Cohere Embeddings reproducibility}

In the next figure the implementation is presented:

\begin{figure}[htp]
    \centering
    \includegraphics[width=10cm]{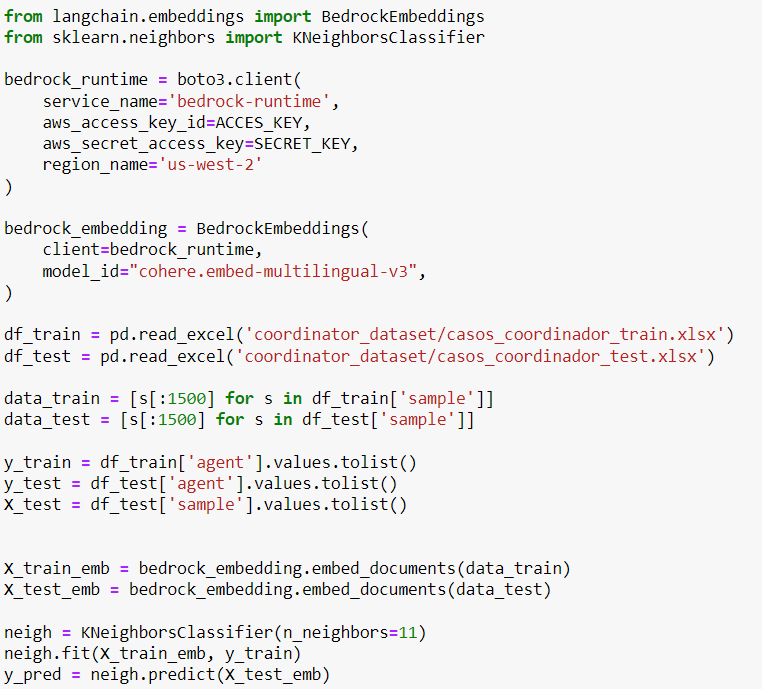}
    \caption{KNN similarity approach: Cohere Embeddings}
    \label{fig:knn2}
\end{figure}

\newpage

\subsection{SVM Based Classifier: Titan Embeddings reproducibility}

In the next figure the implementation is presented:

\begin{figure}[htp]
    \centering
    \includegraphics[width=9cm]{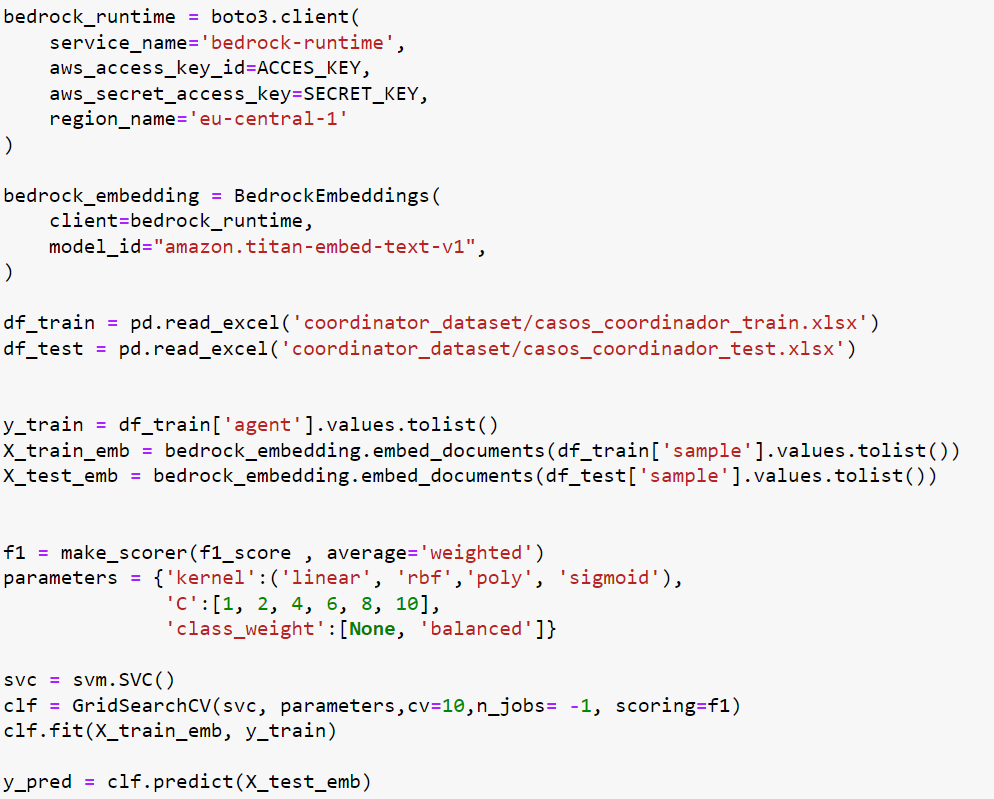}
    \caption{SVM cluster geometry approach: Titan Embeddings}
    \label{fig:svm1}
\end{figure}

\subsection{SVM Based Classifier: Cohere Embeddings reproducibility}

In the next figure the implementation is presented:

\begin{figure}[htp]
    \centering
    \includegraphics[width=9cm]{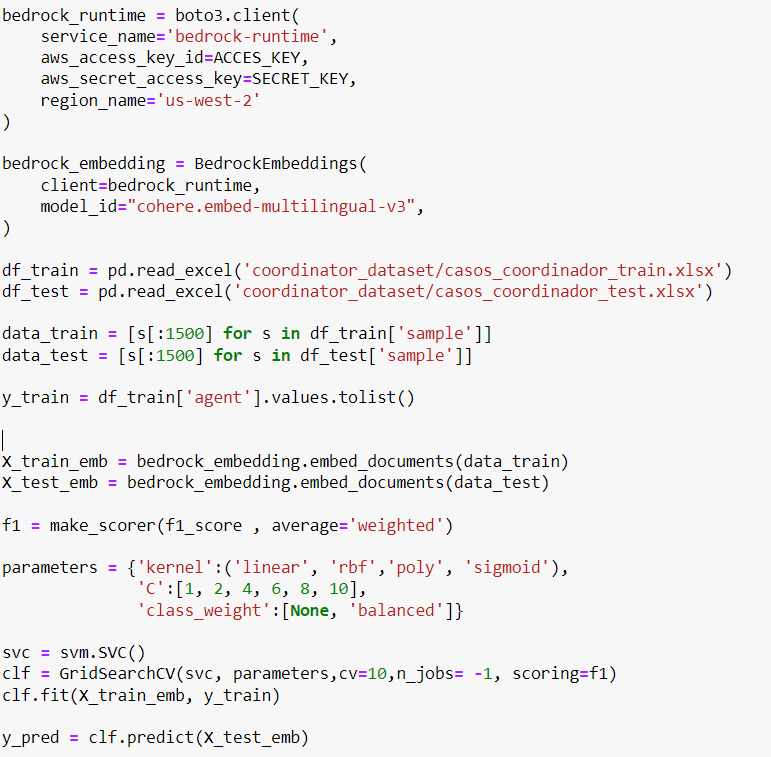}
    \caption{SVM cluster geometry approach: COHERE Embeddings}
    \label{fig:svm2}
\end{figure}

\newpage
\subsection{ANN approach: Titan/Cohere Embeddings reproducibility}

In the next figure the model implementation is presented:

\begin{figure}[htp]
    \centering
    \includegraphics[width=16cm]{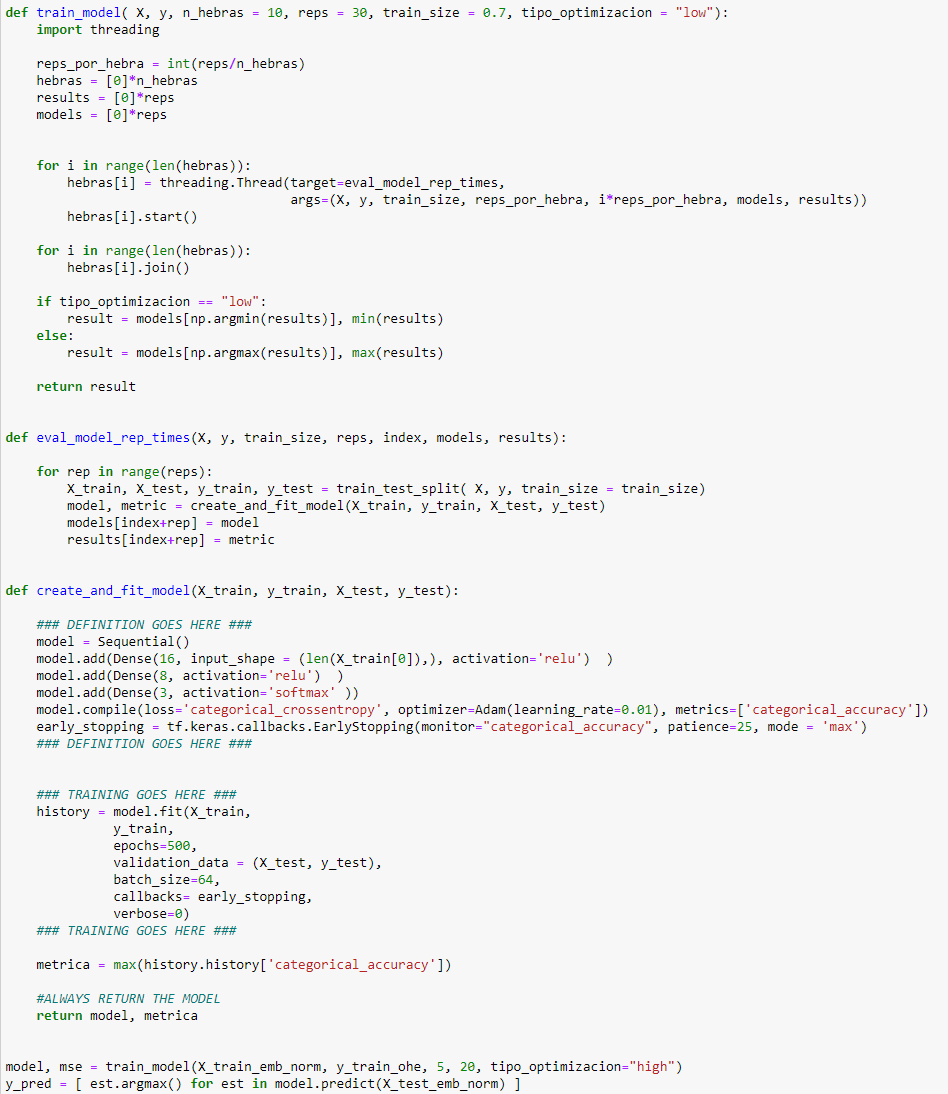}
    \caption{ANN approach: Titan/Cohere Embeddings}
    \label{fig:ann}
\end{figure}
\newpage

In both cases (Titan/Cohere) the same preprocessing has been applied:

\begin{figure}[htp]
    \centering
    \includegraphics[width=12cm]{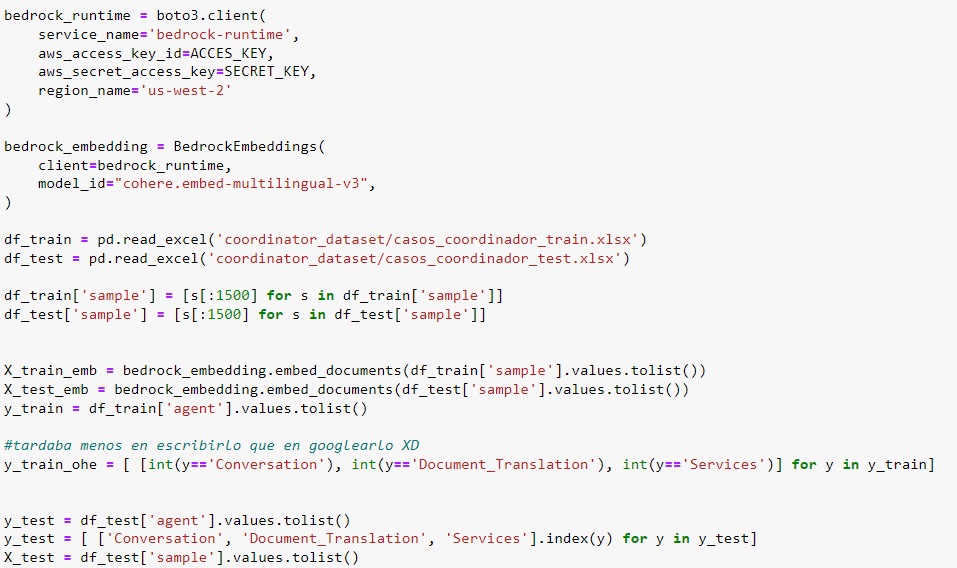}
    \caption{Embeddings preprocessing}
    \label{fig:emb_preprocessing}
\end{figure}

\printbibliography

\end{document}